\documentclass[
    sigconf = true,      
    review = false,       
    screen = true,       
    anonymous = false,    
]{acmart}

\makeatletter
\def\NAT@spacechar{~}
\makeatother

\usepackage{multirow}
\usepackage{mathtools}
\usepackage{pdflscape}
\usepackage{upgreek}
\usepackage[algo2e,ruled,vlined,linesnumbered]{algorithm2e}
\usepackage{todonotes}
\usepackage[nameinlink]{cleveref}

\newcommand{\N}{\ensuremath{\mathbb{N}}} 

\newcommand{\assign}{\leftarrow}

\newcommand{\pmut}{p_{\mathrm{m}}}
\newcommand{\pe}{p_{\mathrm{e}}}
\newcommand{\nind}{n_{\mathrm{ind}}}
\newcommand{\probE}{\mathsf{prob}_{\mathrm{elite}}}

\let\originalleft\left
\let\originalright\right
\renewcommand{\left}{\mathopen{}\mathclose\bgroup\originalleft}
\renewcommand{\right}{\aftergroup\egroup\originalright}

\copyrightyear{2024}
\acmYear{2024}
\setcopyright{rightsretained}
\acmConference[GECCO '24]{2024 Genetic and Evolutionary Computation
 Conference}{July 14--18, 2024}{Melbourne, Australia}
\acmBooktitle{2024 Genetic and Evolutionary Computation Conference
 (GECCO '24), July 10--14, 2024, Melbourne,
 Australia}
 \acmDOI{10.1145/3638529.3654140}
 \acmISBN{979-8-4007-0494-9/24/07}

\acmPrice{15.00}

\begin{document}

\title[Superior Genetic Algorithms for the Target Set Selection Problem]{Superior Genetic Algorithms for the Target Set Selection Problem Based on Power-Law Parameter Choices\\ and Simple Greedy Heuristics}
\titlenote{Author-generated version}

\author{Benjamin Doerr}
\orcid{0000-0002-9786-220X}
\affiliation{%
    \institution{Laboratoire d'Informatique (LIX), CNRS, École Polytechnique,\\ Institut Polytechnique de Paris}
    \city{Palaiseau}
    \country{France}
}

\author{Martin~S. Krejca}
\orcid{0000-0002-1765-1219}
\affiliation{%
    \institution{Laboratoire d'Informatique (LIX), CNRS, École Polytechnique,\\ Institut Polytechnique de Paris}
    \city{Palaiseau}
    \country{France}
}

\author{Nguyen Vu}
\orcid{0000-0000-0000-0000}
\affiliation{%
    \institution{École Polytechnique,\\ Institut Polytechnique de Paris}
    \city{Palaiseau}
    \country{France}
}


\begin{abstract}
    The target set selection problem (TSS) asks for a set of vertices such that an influence spreading process started in these vertices reaches the whole graph.
    The current state of the art for this NP-hard problem are three recently proposed randomized search heuristics, namely a biased random-key genetic algorithm (BRKGA) obtained from extensive parameter tuning, a max-min ant system (MMAS), and a MMAS using Q-learning with a graph convolutional network.

    We show that the BRKGA with two simple modifications and without the costly parameter tuning obtains significantly better results.
    Our first modification is to simply choose all parameters of the BRKGA in each iteration randomly from a power-law distribution. The resulting parameterless BRKGA is already competitive with the tuned BRKGA, as our experiments on the previously used benchmarks show.

    We then add a natural greedy heuristic, namely to  repeatedly discard small-degree vertices that are not necessary for reaching the whole graph. The resulting algorithm consistently outperforms all of the state-of-the-art algorithms.

    Besides providing a superior algorithm for the TSS problem, this work shows that randomized parameter choices and elementary greedy heuristics can give better results than complex algorithms and costly parameter tuning.
\end{abstract}

%
%

\ccsdesc[500]{Mathematics of computing~Combinatorial optimization}

\keywords{Target set selection, combinatorial optimization, biased random-key genetic algorithm, parameter tuning}

\maketitle

\section{Introduction}
\label{sec:introduction}

The \emph{target set selection} problem~\cite{KempeKT15TSSisNPhard} (TSS) is a combinatorial optimization problem that aims to find in a given graph a smallest subset of vertices (the \emph{target set}) such that a diffusion process started on this set eventually reaches all vertices.
Finding such a target set has relevant applications in various domains, especially in viral marketing~\cite{GoldenbergLM01DiffusionModels,MahajanMB90DiffusionModelsSurvey}.
However, as the TSS problem is NP-hard~\cite{KempeKT15TSSisNPhard} and not well approximable~\cite{Chen09TSSisInapproximable}, it is typically solved heuristically.
The state-of-the-art heuristics are the randomized search heuristics that have been proposed in two recent papers~\cite{SerranoB22BRKGA,SanchezSB23QLearning}.

The first of these works, by \citet{SerranoB22BRKGA}, develops a biased random-key genetic algorithm (BRKGA), which encodes potential solutions as a vector of floating-point numbers between~$0$ and~$1$.
Each such vector is translated via a greedy heuristic into a valid target set, that is, a set such that the diffusion process on the instance graph actually reaches all vertices.
The authors tune the parameters of the BRKGA and compare it to the previous state-of-the-art heuristic (called \emph{minimum target set} (MTS)~\cite{Cordasco16MTS}).
On hard instances, they find that the BRKGA almost always finds solutions that are at least~$10$\,\% better than those of the MTS, and this in significantly shorter time, especially for large instances.

The second work, by \citet{SanchezSB23QLearning}, considers two versions of a max-min ant system (MMAS).
One version is a classic MMAS (called MMAS), the other version uses information obtained from a Q-learning approach combined with a graph convolutional network (called MMAS-Learn).
The authors tune the parameters of both versions and then compare them with the BRKGA above~\cite{SerranoB22BRKGA}.
They observe that the two MMAS variants find better solutions than the BRKGA on most of the hard instances.
Especially, the MMAS-Learn is best on large networks.

All of these algorithms have in common that they require considerable effort up front in order to be used to their full effectiveness.
The BRKGA~\cite{SerranoB22BRKGA} has four parameters with a wide range of possible values.
The MMAS variants~\cite{SanchezSB23QLearning} have three parameters each.
Each approach is tuned by the authors via \emph{irace}~\cite{LopezIbanezDLPCBS16irace} before conducting the experiments.
This tuning is computationally very expensive.
For example, both \citet{SerranoB22BRKGA} and \citet{SanchezSB23QLearning} report that they used irace with a budget of~$2\,000$ algorithm runs in order to tune the parameters.
In comparison, when conducting the experimental evaluation, in each of the two papers, the authors perform~$10$ runs for each of the~$27$ instances.
The Q-learning approach combined with the graph convolutional network used for one MMAS variant~\cite{SanchezSB23QLearning} is already a very complex and costly process in itself, for example, requiring to generate training instances for the graph convolutional network.
Overall, since \citet{SerranoB22BRKGA} compared their BRKGA to an existing state-of-the-art TSS heuristic and achieved clearly superior results, this begs the question whether more complex and costly heuristics are necessary in order to achieve good results for the hard TSS problem.

\paragraph{Our contribution.}
We show that the TSS problem can be solved heuristically without the need for complex parameter tuning or other heavy machinery, such as graph convolutional networks.
We propose two modifications to the BRKGA by \citet{SerranoB22BRKGA}, in particular, removing the need for parameter tuning.
Our resulting algorithm outperforms the state-of-the-art algorithms by \citet{SerranoB22BRKGA} and \citet{SanchezSB23QLearning} on almost all hard instances with statistical significance in terms of solution quality (\Cref{tab:significance:fastrevVsRest}).
In terms of computation time, our approach is comparable, while not requiring any offline computations, such as parameter tuning.

Our first modification is to replace the costly parameter tuning by a simple randomized parameter choice in each iteration. For this randomized choice, we use a power-law distribution, cropped and scaled so that it covers exactly the range of parameter values used in~\cite{SerranoB22BRKGA} for the \emph{irace} parameter tuning.

The idea for this type of parameter definition stems from the theoretical work~\cite{DoerrLMN17}, where this approach was used for the first time in discrete evolutionary computation, namely to set the mutation rate of a simple evolutionary algorithm. As shown in~\cite{DoerrLMN17}, this parameter choice gave a drastic speed-up compared to the standard mutation rate $1/n$, and it was only slightly inferior to the instance-dependent optimal mutation rate. The idea to choose parameters in this randomized fashion was quickly taken up in other theoretical works, including works where the power-law choice gave a performance asymptotically faster than with any fixed parameter value~\cite{AntipovBD22} or where several parameters where chosen in this fashion simultaneously~\cite{AntipovBD24}.

Despite this success in theoretical works, power-law distributed parameter choices have not really made it into practical applications of evolutionary algorithms.
Our empirical evaluation shows that there is no reason for this. By simply replacing the parameter choices obtained from expensive tuning by power-law distributed parameters, we obtain a variant of the BRKGA of~\cite{SerranoB22BRKGA} that has a comparable performance (but needed no tuning in the design).

Our second modification is a simple TSS-specific local heuristic that aims at reducing the size of already valid target sets.
It greedily eliminates vertices in the order of increasing vertex degree, always checking whether the result is still valid after the removal.
In addition to adding this modification to the BRKGA, we also combine it with another easy TSS-specific heuristic by \citet{SerranoB22BRKGA}.
In our empirical evaluation with the other approaches, this combination even achieves the best results on some hard instances.

Overall, we observe that easy adjustments can drastically improve the quality for TSS heuristics.
While our experiments are specific to the TSS problem, we believe that the insights carry over to other problems as well.
Coming up with an easy problem-specific heuristics for improving solutions locally is a good choice in general and essentially the same as finding reduction rules, which are very popular, for example, in the PACE challenge~\cite{paceChallenge}.
Furthermore, our recommendation for choosing parameter values on the fly via a power-law distribution is not limited to the TSS problem or the BRKGA, and we invite researchers to try it themselves.

\section{The Target Set Selection Problem}
\label{sec:problem}

The \emph{target set selection} problem (TSS~\cite{KempeKT15TSSisNPhard}) is an NP-hard graph optimization problem.
At its core is a discrete-time diffusion process that determines how vertices in a graph become \emph{active}.
The TSS problem aims to find a minimum cardinality set of initially active vertices \emph{(target set)} such that the diffusion process eventually activates all vertices in the graph.

To make this more precise, for an undirected graph $G = (V, E)$ and for all $v \in V$, let $\Gamma(v) = \{u \in V \mid \{u, v\} \in E\}$ denote the open neighborhood of~$v$, and let $\deg(v) = |\Gamma(v)|$ denote the degree of~$v$.
Given an undirected graph $G = (V, E)$, a threshold function $\theta\colon V \to \N$ such that for all $v \in V$ it holds that $\theta(v) \leq \deg(v)$, as well as a subset $S \subseteq V$ of initially active vertices, the \emph{diffusion process of~$G$ with~$\theta$ and~$S$} is defined as a sequence $(F_t)_{t \in \N}$ over subsets of vertices such that in each iteration vertices whose number of active neighbors is at least the threshold become active as well.
Formally,
\begin{enumerate}
    \item $F_0 = S$ and
    \item $F_{t + 1} = F_t \cup \{v \in V \mid |\Gamma(v) \cap F_t| \geq \theta(v)\}$ for all $t \in \N$.
\end{enumerate}
Let $T = \min \{t \in \N \mid F_{t + 1} = F_t\}$.
Note that since the diffusion process is deterministic, the set of active vertices remains the same for all $t \geq T$.
Since, by definition, the set of active vertices is strictly increasing for all iterations before~$T$, the diffusion process stops after at most $|V| - 1$ iterations.
Let $\sigma_{\theta}(S) \coloneqq F_T$ denote the \emph{finally active set} when starting with set~$S$.
We say that~$S$ is \emph{valid} if and only if $\sigma_{\theta}(S) = V$.

Given an undirected graph $G = (V, E)$ and a threshold function~$\theta$, the TSS problem aims to find a minimum cardinality set $S^* \subseteq V$ such that $\sigma_{\theta}(S^*)$ is valid.

\subsection{Objective Values for the TSS Problem}
\label{sec:problem:fitnessComputation}

In order to solve the TSS problem heuristically, we need to assign an objective value (the \emph{fitness}) to each solution candidate to the problem.
Given a TSS instance $((V, E), \theta)$, each $S \subseteq V$ is a solution candidate.
The fitness of~$S$ is~$|S|$ if~$S$ is valid, and the fitness is $|V| + 1$ otherwise.
Hence, the solution candidates with the smallest fitness are solutions to the original TSS instance.
We note that we only consider algorithms in this article that consider valid solution candidates.
Hence, technically, the fitness of invalid candidates is not required for our purposes.


\section{Existing Heuristics for the Target Set Selection Problem}
\label{sec:knownAlgorithms}

We now present, to the best of our knowledge, the most recent and best performing heuristics for the TSS problem, by \citet{SerranoB22BRKGA} and \citet{SanchezSB23QLearning}.
\citet[Algorithm~$1$]{SerranoB22BRKGA} propose a variant of the \emph{biased random-key genetic algorithm} framework (BRKGA, \Cref{sec:knownAlgorithms:brkga}), which is a genetic algorithm that uses mutation and crossover and that maintains a population of a fixed size.
\citet{SanchezSB23QLearning} consider a \emph{max-min ant system} algorithm (MMAS), with and without heuristic information retrieved via Q-learning from a graph convolutional neural network.
The authors compare these MMAS variants empirically to the BRKGA by \citet{SerranoB22BRKGA} and find that the MMAS with Q-learning finds the best solutions among all three algorithms in~$19$ out of~$27$ cases (including ties), especially for all of the largest networks that were tested.
However, also the MMAS variant without Q-learning as well as the BRKGA perform best in~$11$ and~$10$ cases, respectively (with ties).

All of these algorithms construct valid target sets greedily based via an adaptation of the \emph{maximum-degree heuristic} (MDG)~\cite[Algorithm~$3$]{SerranoB22BRKGA}, see \Cref{sec:knownAlgorithms:mdg} below.
\citet{SerranoB22BRKGA} compared their BRKGA to the MDG alone and observed that the BRKGA found strictly better solutions in all but~$2$ cases, which were ties.


\subsection{Maximum-Degree Heuristic}
\label{sec:knownAlgorithms:mdg}

\begin{algorithm2e}[t]
    \caption{\label{alg:mdg}
        The maximum-degree heuristic (MDG~\cite[Algorithm~$3$]{SerranoB22BRKGA}) that greedily constructs a valid target set.
    }
    \KwIn{graph $G = (V, E)$, threshold function $\theta\colon V \to \N$}
    \SetKwData{cov}{Cov}
    $S \assign \emptyset$\;
    $\cov \assign \emptyset $\;
    \While{$\cov \neq V$}{
        $v^* \assign$ argmax$_{v \in V \setminus \cov} \deg(v)$\; \label{line:mdg:greedyChoice}
        $\cov \assign \sigma_{\theta}(\cov \cup \{v^*\})$\; \label{line:mdg:update}
        $S \assign S \cup \{v^*\}$\;
    }
    \KwOut{$S$}
\end{algorithm2e}

The maximum-degree heuristic (MDG)~\cite[Algorithm~$3$]{SerranoB22BRKGA}, see also \Cref{alg:mdg}, determines for a given graph $G = (V, E)$ and a threshold function~$\theta$ a valid target set $S \subseteq V$, that is, a set such that the diffusion process of~$G$ with~$\theta$ started in~$S$ eventually activates all vertices of~$G$.
To this end, MDG starts with the empty set and greedily adds a vertex of largest degree to its current solution.
After each vertex added, it checks whether the set is already valid.
Once it constructs a valid set, it returns it.

Note that in \cref{line:mdg:update}, we compute the set $\sigma_{\theta}(S \cup \{v^*\})$ of vertices activated by $S \cup \{v^*\}$, exploiting that $\mathsf{Cov} = \sigma_{\theta}(S)$ and $\sigma_{\theta}(S \cup \{v^*\}) = \sigma_{\theta}(\sigma_{\theta}(S) \cup \{v^*\})$~\cite[Proposition~$1$]{SerranoB22BRKGA}.
We perform such an update with a slightly modified breadth-first search.

\subsection{Biased Random-Key Genetic Algorithm}
\label{sec:knownAlgorithms:brkga}

\begin{algorithm2e}[t]
    \caption{\label{alg:brkga}
        The biased random-key genetic algorithm (BRKGA~\cite[Algorithm~$1$]{SerranoB22BRKGA}) for the target set selection problem.
        Each individual is a vector of length~$|V|$, with components in $[0, 1]$.
    }
    \KwIn{graph $G = (V, E)$, threshold function $\theta\colon V \to \N$}
    \KwIn{parameter values $\nind \in \N_{\geq 1}$, $\pe, \pmut \in [0, 1]$ with $\pe + \pmut \leq 1$, $\probE \in [0, 1]$}
    $P \assign$ population of $\nind - 1$ individuals, each generated uniformly at random\;
    $P \assign P \cup \{(0.5)_{i \in V}\}$\;
    \While{termination criterion not met}{
        $P_{\mathrm{e}} \assign$ the best $\lceil \pe \cdot \nind \rceil$ individuals from~$P$\;
        $P_{\mathrm{m}} \assign$ population of $\lceil \pmut \cdot \nind \rceil$ individuals, each generated uniformly at random\;
        $P_c \assign$ population of $\nind - |P_{\mathrm{e}}| - |P_{\mathrm{m}}|$ individuals created by crossover between~$P$ and~$P_{\mathrm{e}}$ with bias~$\probE$\;
        $P \assign P_e \cup P_m \cup P_c$\;
    }
    \KwOut{best solution in~$P$}
\end{algorithm2e}

The biased random-key genetic algorithm (BRKGA) \cite[Algori\-thm~$1$]{SerranoB22BRKGA}, see also \Cref{alg:brkga}, is an elitist genetic algorithm that maintains a population of given fixed size~$\nind$ of \emph{individuals}.
For the TSS problem, each individual is represented as a vector of length~$|V|$, with each component being a floating-point number in $[0, 1]$.
The initial population contains only random individuals as well as one whose values are all~$0.5$.
In each iteration, the BRKGA selects the best solutions from its population, resulting in the \emph{elitist} population.
Then, it creates a population of new random individuals and another population via crossover between the elitist population and the entire population from the previous iteration.
The union of these two populations as well as the elitist population make up the population for the next iteration.

\paragraph{Decoding an individual.}
Following the approach by \citet{SerranoB22BRKGA}, an individual $(w_i)_{i \in V}$ is decoded into a set $S \subseteq V$ that is valid for the TSS problem by following the same approach as MDG but guided by~$w$ instead of the vertex degrees.
More specifically, the decoding performs \Cref{alg:mdg} but replaces \cref{line:mdg:greedyChoice} by
\begin{align*}
    v^* \assign \mathrm{argmax}_{v \in V \setminus \mathrm{Cov}}\ w_v \cdot \deg(v);
\end{align*}

\paragraph{Quality of an individual.}
The quality of an individual is determined by decoding it into a vertex set~$S$ and then determining the objective value of~$S$ as explained in \Cref{sec:problem:fitnessComputation}.
Note that the TSS problem is a minimization problem.
Hence, the term \emph{best} in \Cref{alg:brkga} refers to individuals with the smallest quality-value among the population (breaking ties arbitrarily).

\paragraph{Crossover.}
The crossover operation picks one individual $x \in P$ and one individual $y \in P_{\mathrm{e}}$, each uniformly at random.
It then creates a new individual~$z$ by choosing for each component a value of one of the two parents, with a probability bias of~$\probE$ toward~$y$.
That is, for all $v \in V$, it holds independently with probability~$\probE$ that $z_v = y_v$, and it holds $z_v = x_v$ otherwise.

\section{Our Improvements to the Existing Heuristics}
\label{sec:improvedBRKGA}

We propose two independent modifications to the BRKGA described in \Cref{sec:knownAlgorithms}, namely:
\begin{enumerate}
    \item On-the-fly parameter choices based on a power-law distribution (\Cref{sec:improvedBRKGA:parameterTuning}).
    \item The \emph{minimum-degree heuristic} (reverseMDG, \Cref{alg:reverseMDG}), which is given a valid solution candidate for the TSS problem and greedily aims to improve it.
\end{enumerate}

\subsection{On-the-Fly Parameter Tuning via Power-Law Random Choices}
\label{sec:improvedBRKGA:parameterTuning}

To avoid the costly tuning of the parameters of the BRKGA, we propose to choose the parameter values during the run, namely randomly from a power-law distribution.
Naturally, this only concerns parameters that can be adjusted meaningfully during the run. For the BRKGA, these are all parameters except the population size, which we choose following the recommendation by \citet{SerranoB22BRKGA}.

Power-law distributed random parameter values were proposed first (in evolutionary computation with discrete search spaces) in the theoretical work~\cite{DoerrLMN17}, where under the name \emph{fast GA} it was suggested to use bit-wise mutation with a random mutation rate, sampled anew for each application of the mutation operator from a power-law distribution.
This idea was quickly taken up in other theoretical works and extended to other parameters, to more than one parameter, and to multi-objective optimization~\cite{FriedrichQW18,FriedrichGQW18,WuQT18,AntipovD20ppsn,QuinzanGWF21,AntipovBD24,DoerrZ21aaai,CorusOY21tec,AntipovBD22,DangELQ22,DoerrGI24,DoerrR23,DoerrQ23tec}.
All these work showed that a power-law parameter choice can significantly beat the standard parameter choice, or even any static parameter choice, and this in an essentially parameterless manner (formally speaking, the power-law exponent~$\beta$ is a parameter, but already the first work~\cite{DoerrLMN17} detected that it has little influence on the performance and that taking $\beta=1.5$ is a reasonable choice). Surprisingly, these promising theoretical results never found significant applications in practice.

\paragraph{Formal definition of capped power-laws.}
In~\cite{DoerrLMN17}, the following type of capped power-law was used. Let $\beta > 1$ be the (negative) power-law exponent and $[1..r] \coloneqq [1, r] \cap \N$, $r \in \N$, be the range of the distribution.
Then a random variable $X$ follows a power-law with exponent~$\beta$ and range $[1..r]$ if for all $k \in [1..r]$,
\[
\Pr[X = k] = \frac{k^{-\beta}}{C_r(\beta)},
\]
where $C_r = \sum_{k=1}^r k^{-\beta}$.


\citet{DoerrLMN17} report that the choice of~$\beta$ does not have a big impact on the run time, and they recommend a choice of $\beta = 1.5$. The range of the power-law is usually determined by the application. In~\cite{DoerrLMN17}, a random mutation rate for mutation of bit-strings of length~$n$ was sought, so with the aim of avoiding mutation rates larger than $1/2$, the number $r$ was chosen as $r=\frac n2$, a number $\alpha$ was sampled from a power-law with range $[1..r]$, and $\alpha/n$ was used as mutation rate.
These considerations turn the power-law distribution effectively into a parameter-less distribution.


It is clear that the range of the power-law distribution can be adjusted to arbitrary finite sets~$A$ by choosing an appropriate mapping between~$A$ and~$[1..r]$.


\subsubsection{Applying Power-Law Tuning to the BRKGA}
\label{sec:improvedBRKGA:parameterTuning:brkga}

\begin{table}
    \caption{\label{tab:powerLawChoices}
        The power-law distributions we employ when choosing the parameters of the BRKGA (\Cref{alg:brkga}) according to the method described in \Cref{sec:improvedBRKGA:parameterTuning}.
        The column \emph{Support} denotes the values that the power-law distribution takes.
        All distributions choose a power-law exponent of~$1.5$.
        The conversion map shows how we transform the values from \emph{Support} such that they result in a power-law over the values in \emph{Resulting range}, which are the ones chosen by \citet[Table~$1$]{SerranoB22BRKGA}.
        The value~$x$ represents a random number drawn from the respective power-law distribution.
    }
    \begin{tabular}{llll}
        Parameter & Support & Conversion map & Resulting range\\
        \toprule
        $\pe$ & $[1 .. 15]$ & $0.1 + 0.01(15 - x)$ & $\{0.24, 0.23, \ldots, 0.1\}$\\
        $\pmut$ & $[1 .. 20]$ &  $0.1+0.01x$ & $\{0.11, 0.12, \ldots, 0.3\}$\\
        $\probE$ & $[1 .. 30]$ & $0.5+0.01x$ & $\{0.51, 0.52, \ldots, 0.8\}$
    \end{tabular}
\end{table}

Out of the four parameters of the BRKGA (\Cref{alg:brkga}), we choose all but~$\nind$ via a power-law distribution.\footnote{Technically, the BRKGA by \citet{SerranoB22BRKGA} has one more parameter, which is binary and affects a single individual in the initial population. Since it does not make sense to adjust such a parameter dynamically, we go with the choice that \citet{SerranoB22BRKGA} propose (i.e., $\mathsf{seed} = \mathsf{true}$). \Cref{alg:brkga} already reflects this choice.}

For the three remaining parameters, we pick a power-law over the range of values that \citet[Table~$1$]{SerranoB22BRKGA} chose for their parameter tuning.
We apply a suitable transformation, as reported in \Cref{tab:powerLawChoices}.
We note that the ranges of~$\pe$ and~$\pmut$ do not overlap and are such that the constraint $\pe + \pmut \leq 1$ is always satisfied.
However, we only make this choice because we choose for each parameter the same range as \citet[Table~$1$]{SerranoB22BRKGA}.
The constraint $\pe + \pmut \leq 1$ can still be satisfied if the ranges of~$\pe$ and~$\pmut$ would overlap.
In such a case, one would first sample one of the two values, for example~$\pe$, and then sample the other value via rejection sampling.
Alternatively, the support of the power-law mutation for the second could be adjusted.

When transforming the values of the power-law into the intended parameter range, it is not very important where the highest probability mass lies, as the ranges are all rather small.
For example, for~$\pe$, the highest mass is on the value that \citet{SerranoB22BRKGA} determined as best via the parameter tuner irace~\cite{LopezIbanezDLPCBS16irace}.
However, for~$\probE$, the best value reported by \citet{SerranoB22BRKGA} is~$0.69$, which receives a comparably low probability in the power-law we chose.

We recall that we draw a value for each of these three parameters once at the beginning of the \textsf{while}-loop of the BRKGA, that is, once each iteration.

\begin{table*}
    \caption{\label{tab:bestFitness:newHeuristics}
        The best TSS objective values achieved for all of the algorithms that we ran on the specified networks, as described in \Cref{sec:experiments:setup}.
        The average is over~$10$ independent runs per algorithm.
        Bold numbers indicate the best value among all algorithms in this table and in \Cref{tab:bestFitness:existingHeuristics}.
        The columns~$|V|$ and~$|E|$ denote the number of vertices and edges, respectively, of the networks.
        See \Cref{sec:experiments:results:runTime} for more details.
    }
    \begin{tabular}{lrr@{\hspace*{0.75 em}}rr r@{\hspace*{0.75 em}}rr@{\hspace*{0.75 em}}rr@{\hspace*{0.75 em}}rr@{\hspace*{0.75 em}}rr}
    \multirow{2}{*}{Network} & \multirow{2}{*}{$|V|$} & \multirow{2}{*}{$|E|$} & \multicolumn{2}{c}{BRKGA} & \multicolumn{1}{r}{\multirow{2}{*}{MDG+rev}} & \multicolumn{2}{c}{BRKGA+rev} & \multicolumn{2}{c}{fastBRKGA}    & \multicolumn{2}{c}{fastBRKGA+rev} \\ \cmidrule(lr){4-5}\cmidrule(lr){7-8}\cmidrule(lr){9-10}\cmidrule(lr){11-12}
                             &                        &                        & Best         & Avg.       &                                              & Best             & Avg.       & Best         & Avg.              & Best            & Avg.            \\ \toprule
    Dolphins                 & 62                     & 159                    & \textbf{6}   & 6.0        & 7                                            & \textbf{6}       & 6.0        & \textbf{6}   & 6.0               & \textbf{6}      & 6.0             \\
    Football                 & 115                    & 613                    & \textbf{22}  & 23.4       & 28                                           & \textbf{22}      & 23.4       & \textbf{22}  & 23.5              & \textbf{22}     & 23.5            \\
    Karate                   & 34                     & 78                     & \textbf{3}   & 3.0        & \textbf{3}                                   & \textbf{3}       & 3.0        & \textbf{3}   & 3.0               & \textbf{3}      & 3.0             \\
    Jazz                     & 198                    & 2\,742                 & \textbf{20}  & 21.1       & 24                                           & \textbf{20}      & 21.1       & \textbf{20}  & 21.1              & \textbf{20}     & 21.1            \\
    CA-AstroPh               & 18\,772                & 198\,050               & 1\,428       & 1\,438.0   & 1\,381                                       & \textbf{1\,375}  & 1\,384.9   & 1\,427       & 1\,438.4          & \textbf{1\,375} & 1\,385.6        \\
    CA-GrQc                  & 5\,242                 & 14\,484                & 928          & 930.7      & \textbf{889}                                 & 891              & 895.7      & 924          & 930.6             & 892             & 897.0           \\
    CA-HepPh                 & 12\,008                & 118\,489               & 1\,343       & 1\,347.8   & \textbf{1\,257}                              & 1\,280           & 1\,286.0   & 1\,338       & 1\,349.2          & 1\,278          & 1\,285.7        \\
    CA-HepTh                 & 9\,877                 & 25\,973                & 1\,234       & 1\,242.4   & \textbf{1\,154}                              & 1\,155           & 1\,160.2   & 1\,237       & 1\,242.4          & 1\,157          & 1\,160.6        \\
    CA-CondMat               & 23\,133                & 93\,439                & 2\,563       & 2\,592.8   & \textbf{2\,326}                              & 2\,350           & 2\,360.4   & 2\,580       & 2\,599.5          & 2\,354          & 2\,364.1        \\
    Email-Enron              & 36\,692                & 183\,831               & 2\,670       & 2\,679.1   & 2\,676                                       & 2\,635           & 2\,643.5   & 2\,664       & 2\,671.7          & \textbf{2\,633} & 2\,639.1        \\
    ego-facebook             & 4\,039                 & 88\,234                & 466          & 470.5      & 477                                          & 463              & 464.7      & 467          & 473.2             & \textbf{460}    & 466.7           \\
    socfb-Brandeis99         & 3\,898                 & 137\,567               & \textbf{320} & 337.4      & 356                                          & \textbf{320}     & 336.2      & 321          & 332.4             & \textbf{320}    & 331.5           \\
    socfb-nips-ego           & 2\,888                 & 2\,981                 & \textbf{10}  & 10.0       & \textbf{10}                                  & \textbf{10}      & 10.0       & \textbf{10}  & 10.0              & \textbf{10}     & 10.0            \\
    socfb-Mich67             & 3\,748                 & 81\,903                & 166          & 168.1      & 177                                          & 165              & 168.0      & \textbf{164} & 167.9             & \textbf{164}    & 167.9           \\
    soc-gplus                & 23\,628                & 39\,194                & 62           & 63.0       & \textbf{61}                                  & \textbf{61}      & 61.0       & 62           & 62.8              & \textbf{61}     & 61.1            \\
    musae\_git               & 37\,700                & 289\,003               & 173          & 178.4      & 190                                          & 172              & 178.1      & 172          & 178.3             & \textbf{171}    & 178.1           \\
    loc-gowalla\_edges       & 196\,591               & 950\,327               & 5\,450       & 5\,465.2   & 4\,780                                       & \textbf{4\,760}  & 4\,777.2   & 5\,436       & 5\,464.6          & 4762            & 4\,778.9        \\
    gemsec\_facebook\_artist & 50\,515                & 819\,090               & 648          & 669.2      & 688                                          & 631              & 649.8      & 624          & 671.1             & \textbf{610}    & 650.6           \\
    deezer\_HR               & 54\,573                & 498\,202               & 2\,046       & 2\,092.1   & 1\,895                                       & 1\,874           & 1\,904.5   & 2\,057       & 2\,096.5          & \textbf{1\,873} & 1\,892.2        \\
    com-dblp                 & 317\,080               & 1\,049\,866            & 36\,875      & 36\,917.0  & \textbf{29\,114}                             & 29\,170          & 29\,215.8  & 36\,832      & 36\,907.8         & 29\,212         & 29\,235.5       \\
    Amazon0302               & 262\,111               & 899\,792               & 35\,579      & 35\,567.3  & \textbf{26\,618}                             & 26\,665          & 26\,642.0  & 35\,576      & 35\,563.9         & 26\,667         & 26\,636.4       \\
    Amazon0312               & 400\,727               & 2\,349\,869            & 31\,051      & 31\,075.2  & 23\,529                                      & \textbf{23\,523} & 23\,549.6  & 31\,045      & 31\,068.2         & 23\,542         & 23\,559.4       \\
    Amazon0505               & 410\,236               & 2\,439\,437            & 31\,763      & 31\,778.5  & \textbf{24\,115}                             & \textbf{24\,115} & 24\,134.6  & 31\,751      & 31\,776.9         & 24\,122         & 24\,135.0       \\
    Amazon0601               & 403\,394               & 2\,443\,408            & 31\,393      & 31\,412.3  & 23\,759                                      & \textbf{23\,757} & 23\,777.1  & 31\,390      & 31\,403.0         & 23\,758         & 23\,777.6
    \end{tabular}
\end{table*}

\subsection{ReverseMDG}
\label{sec:improvedBRKGA:reverseMDG}

\begin{algorithm2e}[t]
    \caption{\label{alg:reverseMDG}
        The minimum-degree heuristic (reverseMDG) that is given a valid solution candidate for the target set selection problem and greedily tries to eliminate vertices of the lowest degree.
    }
    \KwIn{graph~$G$, threshold function~$\theta$, and $S \subseteq V$ such that $\sigma_{\theta}(S) = V$}
    $V_{\leq} \assign $ vertices in~$V$ sorted by ascending vertex degree\;
    $C \assign S$\;
    \For{$v \in V_{\leq}$}{
        \If{$v \in C$}{
            $C' \assign C \setminus \{v\}$\;
            \lIf{$\sigma_{\theta}(C') = V$}{%
                $C \assign C'$%
            }
        }
    }
    \KwOut{$C$}
\end{algorithm2e}

The minimum-degree heuristic (reverseMDG, \Cref{alg:reverseMDG}) is given a valid TSS solution candidate~$S$ and greedily tries to eliminate vertices from~$S$, starting with those of the lowest degree.
In our BRKGA, we apply the heuristic after obtaining target sets from the MDG heuristic (\Cref{alg:mdg}).

Since the TSS problem is NP-hard, a heuristic solution candidate, such as one constructed by MDG, may not be optimal because~(1) it may not contain vertices that are required for an optimal solution or~(2) it may not have minimum size.
While reverseMDG does not help with~(1), it helps with~(2).
Since vertices with a high degree can potentially activate more vertices once they are active themselves, reverseMDG tries to remove vertices with a low degree first.
Although this does not even guarantee a locally optimal solution, we see in \Cref{sec:experiments} that reverseMDG is already very powerful.
We note that, mostly due to its simplicity, reverseMDG is also comparably cheap to compute:
It only needs to compute the finally active set for a given set at most~$|S|$ times, which is typically not too large, as we are considering a minimization problem.

\begin{table}
    \caption{\label{tab:bestFitness:existingHeuristics}
        The empirical results from \citet[Table~$2$]{SanchezSB23QLearning} for solving the TSS instances as explained in \Cref{sec:experiments:setup}.
        The numbers are the objective values of the best solution found after a certain time budget.
        The average is over~$10$ independent runs per network.
        Numbers in bold are the best value among all of the algorithms in this table and in \Cref{tab:bestFitness:newHeuristics}.
        We note that we strongly believe that \citet[Table~$2$]{SanchezSB23QLearning} mixed up the labels of the rows for \texttt{Dolphins} and for \texttt{Karate}, as the number of vertices and edges do not match the actual values of these networks.
        Below, we corrected this.
        See \Cref{sec:experiments:results:runTime} for more details.
    }
    \begin{tabular}{l@{\hspace*{1 em}}rr@{\hspace*{1.4 em}}rr}
        \multirow{2}{*}{Network} & \multicolumn{2}{c}{MMAS} & \multicolumn{2}{c}{MMAS-Learn} \\ \cmidrule(lr){2-3}\cmidrule(lr){4-5}
                                 & Best        & Avg.       & Best        & Avg.             \\ \toprule
        Dolphins                 & \textbf{6}  & 6.0        & \textbf{6}  & 6.0              \\
        Football                 & 23          & 23.0       & \textbf{22} & 23.0             \\
        Karate                   & \textbf{3}  & 3.0        & \textbf{3}  & 3.0              \\
        Jazz                     & \textbf{20} & 20.0       & \textbf{20} & 20.0             \\
        CA-AstroPh               & 1\,405      & 1\,412.5   & 1\,405      & 1\,413.0         \\
        CA-GrQc                  & 898         & 900.1      & 897         & 899.4            \\
        CA-HepPh                 & 1\,289      & 1\,297.2   & 1\,289      & 1\,298.4         \\
        CA-HepTh                 & 1\,179      & 1\,186.2   & 1\,182      & 1\,189.2         \\
        CA-CondMat               & 2\,416      & 2\,422.3   & 2\,419      & 2\,428.0         \\
        Email-Enron              & 2\,679      & 2\,686.0   & 2\,692      & 2\,699.4         \\
        ego-facebook             & 478         & 482.8      & 478         & 481.9            \\
        socfb-Brandeis99         & 365         & 369.2      & 368         & 370.2            \\
        socfb-nips-ego           & \textbf{10} & 10.0       & \textbf{10} & 10.0             \\
        socfb-Mich67             & 177         & 179.3      & 177         & 179.3            \\
        soc-gplus                & \textbf{61} & 61.5       & \textbf{61} & 61.9             \\
        musae\_git               & 202         & 205.9      & 199         & 206.4            \\
        loc-gowalla\_edges       & 5\,180      & 5\,195.1   & 5\,155      & 5\,177.6         \\
        gemsec\_facebook\_artist & 726         & 744.7      & 735         & 748.4            \\
        deezer\_HR               & 2\,231      & 2\,247.0   & 2\,240      & 2\,255.9         \\
        com-dblp                 & 32\,981     & 33\,016.9  & 32\,364     & 32\,397.7        \\
        Amazon0302               & 30\,291     & 30\,334.7  & 29\,553     & 29\,618.0        \\
        Amazon0312               & 26\,280     & 26\,317.8  & 26\,186     & 26\,201.2        \\
        Amazon0505               & 26\,945     & 27\,000.9  & 26\,801     & 26\,871.2        \\
        Amazon0601               & 26\,665     & 26\,708.3  & 26\,511     & 26\,568.0        \\
    \end{tabular}
\end{table}

\begin{table}
    \caption{\label{tab:significance:tunedVsFast}
        The statistical significance of the best solution found of the \emph{fastBRKGA compared to the (tuned) BRKGA} by \citet{SerranoB22BRKGA}.
        The reported $p$-values are the result of a Mann--Whitney $U$ test based on the same~$10$ independent runs per algorithm per network as the ones in \Cref{tab:bestFitness:newHeuristics}.
        Values of statistical significance (that is, a $p$-value of at most~$0.05$) are highlighted in bold.
        See \Cref{sec:experiments:results:significance} for more details.
    }
    \begin{tabular}{l r}
        Network                  & $p$-value     \\ \toprule
        Dolphins                 & 1.00          \\
        Football                 & 0.77          \\
        Karate                   & 1.00          \\
        Jazz                     & 0.56          \\
        CA-AstroPh               & 0.82          \\
        CA-GrQc                  & 0.91          \\
        CA-HepPh                 & 0.70          \\
        CA-HepTh                 & 0.79          \\
        CA-CondMat               & 0.57          \\
        Email-Enron              & \textbf{0.03} \\
        ego-facebook             & 0.10          \\
        socfb-Brandeis99         & 0.27          \\
        socfb-nips-ego           & 1.00          \\
        socfb-Mich67             & 0.97          \\
        soc-gplus                & 0.36          \\
        musae\_git               & 1.00          \\
        loc-gowalla\_edges       & 1.00          \\
        gemsec\_facebook\_artist & 0.60          \\
        deezer\_HR               & 0.60          \\
        com-dblp                 & 0.62          \\
        Amazon0302               & 0.33          \\
        Amazon0312               & 0.24          \\
        Amazon0505               & 1.00          \\
        Amazon0601               & \textbf{0.05} \\
    \end{tabular}
\end{table}

\section{Empirical Evaluation}
\label{sec:experiments}

We compare the existing heuristics for the TSS problem discussed in \Cref{sec:knownAlgorithms} to a select choice of these algorithms modified by our approaches discussed in \Cref{sec:improvedBRKGA}.
More specifically, we compare the BRKGA from \citet{SerranoB22BRKGA} as well as the MMAS with and without Q-learning from \citet{SanchezSB23QLearning} to the BRKGA with power-law tuning (\Cref{sec:improvedBRKGA:parameterTuning:brkga}), the BRKGA with power-law tuning and with reverseMDG, and the MDG with reverseMDG.
We compare these algorithms primarily with respect to the best solution that they found after a fixed time budget.
However, we also report on the run time.
Our code is available on GitHub~\cite{data}.

We chose to apply our modifications to the BRKGA and the MDG, because they are simple algorithms, and we are interested to see to what extent our modifications improve such simple approaches, especially compared to more sophisticated approaches such as MMAS with Q-learning.
For modifications that employ the reverseMDG, we add the suffix \emph{+rev} to the algorithm name.
For modifications that employ the power-law parameter tuning, we add the prefix \emph{fast} to the algorithm name, based on the naming convention by \citet{DoerrLMN17}.

\subsection{Experimental Setup}
\label{sec:experiments:setup}
We consider the same experimental setup as \citet{SerranoB22BRKGA} and \citet{SanchezSB23QLearning}.
To this end, we consider $24$ out of the~$27$ social networks that the previous studies considered that are part of the Stanford Network Analysis Project~\cite{LeskovecK14DataSets}.
We note that we exclude the network \texttt{com-youtube} from this data set, due to its size---being at least~$2.5$ times larger than the largest of the remaining networks, in terms of vertices.
The two networks \texttt{ncstrlwg2} and \texttt{actors-data} from the previous studies are excluded because we could not obtain them.

For each graph~$(V, E)$ above, we consider a TSS instance where the threshold of each vertex is half its degree, that is, for each $v \in V$, it holds that $\theta(v) = \lceil \deg(v) / 2 \rceil$.

We run the original BRKGA from \citet{SerranoB22BRKGA} with the best parameters determined by the authors.
In addition, we run our modifications MDG+rev, BRKGA+rev, fastBRKGA, and fastBRKGA+rev.
Since the BRKGA+rev does not tune its parameters automatically, we choose the best parameters determined by \citet{SerranoB22BRKGA}.
For all versions of the BRKGA, we choose the population size $\nind = 46$ recommended by \citet{SerranoB22BRKGA}.
We execute~$10$ independent runs per algorithm and TSS instance.
For each graph $(V, E)$, each run has a run time budget of $\max \{100, |V|/100\}$ seconds.
We stop a run prematurely if it finds an optimal solution.
During each run, we log the objective value of the best solution found so far.
For the BRKGA and the fastBRKGA, we also log the run times.

All of our modifications were implemented in C++, since \citet{SerranoB22BRKGA} also used C++ for the BRKGA.
Our experiments were run on a machine with~$2$ Intel\textsuperscript{\textregistered} Xeon\textsuperscript{\textregistered} Platinum $8362$ CPUs @~$2.80$\,GHz ($32$~cores; $64$~threads) with $2$\,TB of RAM.

\subsection{Results}
\label{sec:experiments:results}

We discuss our results mostly with respect to the quality of the best solution found.
We do so first in absolute terms (\Cref{sec:experiments:results:solutionQuality}) and then in terms of statistical significance (\Cref{sec:experiments:results:significance}).
Last, we briefly discuss the run time, since we compare our results to some algorithms that we did not run ourselves (\Cref{sec:experiments:results:runTime}).

\subsubsection{Comparison of the Best Objective Value}
\label{sec:experiments:results:solutionQuality}

\Cref{tab:bestFitness:newHeuristics} shows our results in terms of objective value.
\Cref{tab:bestFitness:existingHeuristics} shows the results obtained by \citet{SanchezSB23QLearning} for the same setting; that is, we did not run experiments for the results in \Cref{tab:bestFitness:existingHeuristics}.

First, we observe that all bold entries for the existing heuristics (that is, BRKGA, MMAS, and MMAS-learn) are almost always matched by all of our modifications of the BRKGA.
The only two exceptions are that the fastBRKGA reports for \texttt{soc-gplus} and for \texttt{socfb-Brandeis99} values that are worse by~$1$ than the overall best value.
For all of these networks, the average is for all algorithms also very close to the best value, indicating that these are not particularly hard instances.
Still, our modification MDG+rev does not match the best value in~$3$ cases, showing that a purely greedy strategy is insufficient.
In total, these cases cover~$6$ out of the~$24$ networks.

For the remaining~$18$ networks on which at least one of our modifications is solely the best algorithm, we get a diverse picture.
Surprisingly, the deterministic greedy heuristic MDG+rev obtains the best value in~$7$ of these~$18$ cases.
This implies for these instances that the randomness used by any of the other algorithms is worse than a simple one-time greedy approach.
This is even the case for the fastBRKGA+rev, which employs a modified version of MDG+rev in order to create valid solutions.
This suggests that it would be generally better to add the solution computed by MDG+rev to the initial population of every other algorithm, which would remain in the population, due to the elitist nature of the algorithms.

Focusing on our modifications of the BRKGA on the $18 - 7 = 11$ remaining networks, we see that the fastBRKGA performs worst in terms of number of best solutions found.
However, when compared to the original BRKGA, which was tuned in advance, the results are very similar.
This shows that the power-law parameter tuning is actually comparable to the offline parameter tuning of the BRKGA via irace.
We consider this to be a success for the power-law tuning.

The two remaining algorithms are the BRKGA+rev and the fastBRKGA+rev, which only differ on an algorithmic level with respect to whether their parameters were tuned before the optimization or chosen during the optimization.
We observe that the fastBRKGA+rev performs better on smaller networks, whereas the BRKGA+rev performs better on larger networks.
However, for each of these networks, the differences in the best solution found are rather close.
Any of these two modified algorithms is better than the state of the art, presented in \Cref{tab:bestFitness:existingHeuristics}.
Overall, this shows that the reverseMDG heuristic, which is problem-dependent, has an important impact on the overall performance.
Furthermore, the simple power-law parameter tuning is comparable to the complex offline parameter tuning in terms of the solution quality the respective variants produce.

\subsubsection{Statistical Significance}
\label{sec:experiments:results:significance}

\Cref{tab:significance:fastrevVsRest} shows a comparison in terms of statistical significance of the best solution found of the fastBRGKA+rev to the existing TSS heuristics as well as our modifications.
We notice that our observations from \Cref{sec:experiments:results:solutionQuality} are confirmed.
For most of the networks, besides the easy ones, the fastBRKGA+rev is significantly better; in many cases with a very small $p$-value.
This advantage holds also against the fastBRKGA, but it does not exist against the BRKGA+rev.
This shows that the reverseMDG heuristic, albeit simple, has a huge impact on the performance of the algorithms.

\Cref{tab:significance:tunedVsFast} as well as the last column of \Cref{tab:significance:fastrevVsRest} compare the BRKGA variants whose parameters were tuned offline (by \citet{SerranoB22BRKGA}) against the respective variants who use the on-the-fly power-law parameter tuning.
We see no statistical significance in either case, except for two cases in \Cref{tab:significance:tunedVsFast}, where the fastBRKGA is significantly better than the (tuned) BRKGA.
This highlights that an on-the-fly parameter choice following a power-law is comparable to costly offline parameter tuning in this setting.

\begin{table}[h]
    \caption{\label{tab:runTime}
        The average run time in seconds of all of the algorithms for which we have run time information.
        We note that we did not run MMAS and MMAS-Learn (MMAS-L below) ourselves---their respective columns are from \citet[Table~$2$]{SanchezSB23QLearning}.
        For the BRKGA and the fastBRKGA, the information is from the runs from \Cref{tab:bestFitness:newHeuristics}.
        The sizes of the networks are mentioned in \Cref{tab:bestFitness:newHeuristics}.
        Bold entries denote the minimum value of a row.
        We strongly believe that \citet{SanchezSB23QLearning} mixed up the labels of the rows \texttt{Dolphins} and \texttt{Karate}, which we correct below.
    }
    \begin{tabular}{l r r r r}
    Network                  & BRKGA           & MMAS            & MMAS-L            & fBRKGA          \\ \toprule
    Dolphins                 & \textbf{< 0.01} & \textbf{< 0.01} & \textbf{< 0.01}   & \textbf{< 0.01} \\
    Football                 & \textbf{10.2}   & 22.0            & 30.3              & 21.3            \\
    Karate                   & \textbf{< 0.01} & \textbf{< 0.01} & \textbf{< 0.01}   & \textbf{< 0.01} \\
    Jazz                     & 12.3            & \textbf{4.7}    & 12.9              & 5.8             \\
    CA-AstroPh               & \textbf{162.2}  & 173.1           & 172.6             & 176.3           \\
    CA-GrQc                  & \textbf{51.3}   & 64.2            & 72.3              & 52.1            \\
    CA-HepPh                 & 102.90          & 109.2           & 110.5             & \textbf{84.9}   \\
    CA-HepTh                 & 88.20           & 84.6            & 82.5              & \textbf{82.3}   \\
    CA-CondMat               & 225.80          & \textbf{220.4}  & 222.8             & 226.2           \\
    Email-Enron              & 328.9           & \textbf{287.4}  & 299.6             & 334.0           \\
    ego-facebook             & \textbf{51.3}   & 65.7            & 68.1              & 54.9            \\
    socfb-Brandeis99         & \textbf{53.9}   & 68.1            & 59.3              & 56.0            \\
    socfb-nips-ego           & \textbf{< 0.01} & \textbf{< 0.01} & \textbf{< 0.01}   & \textbf{< 0.01} \\
    socfb-Mich67             & \textbf{26.1}   & 68.3            & 67.5              & 34.8            \\
    soc-gplus                & 15.30           & 28.8            & \textbf{10.2}     & 44.2            \\
    musae\_git               & \textbf{169.4}  & 202.9           & 172.7             & 192.9           \\
    loc-gowalla\_edges       & 1\,938.5        & \textbf{728.0}  & 1\,113.5          & 1\,950.5        \\
    gemsec\_facebook-        & 457.3           & 383.6           & \textbf{330.8}    & 445.9           \\
    \hspace*{1 em}\_artist                 &                 &                 &                   &                 \\
    deezer\_HR               & 541.6           & 335.9           & \textbf{283.9}    & 535.3           \\
    com-dblp                 & 3\,059.9        & 1\,749.1        & \textbf{1\,726.5} & 3\,065.1        \\
    Amazon0302               & 2\,527.4        & 1\,768.8        & \textbf{1\,705.4} & 2\,524.4        \\
    Amazon0312               & 3\,820.4        & 1\,747.4        & \textbf{1\,733.1} & 3\,851.3        \\
    Amazon0505               & 3\,993.5        & 1\,746.3        & \textbf{1\,695.6} & 3\,965.8        \\
    Amazon0601               & 3\,862.5        & 1\,726.5        & \textbf{1\,714.0} & 3\,900.3
    \end{tabular}
\end{table}

\begin{table*}
    \caption{\label{tab:significance:fastrevVsRest}
        The statistical significance ($p$-values) of the best solution found of the \emph{fastBRKGA+rev} in comparison to the other shown algorithms.
        The reported $p$-values are the result of a Mann--Whitney $U$ test based on the same~$10$ independent runs per algorithm per network as the ones in \Cref{tab:bestFitness:newHeuristics}.
        For MMAS and MMAS-Learn, since we did not run any experiments, we use for each instance the two data points (best and average) from \Cref{tab:bestFitness:existingHeuristics}.
        Values of statistical significance (that is, with a $p$-value of at most~$0.05$) are highlighted in bold.
        See \Cref{sec:experiments:results:significance} for more details.
    }
    \begin{tabular}{l *{5}{r}}
        Network                  & BRKGA            & MMAS             & MMAS-Learn    & fastBRKGA     & BRKGA+rev \\ \toprule
        Dolphins                 & 1.00             & 1.00             & 1.00          & 1.00          & 1.00      \\
        Football                 & 0.63             & 0.51             & 0.89          & 0.96          & 0.77      \\
        Karate                   & 1.00             & 1.00             & 1.00          & 1.00          & 1.00      \\
        Jazz                     & 0.72             & 0.50             & 0.91          & 0.91          & 0.56      \\
        CA-AstroPh               & \textbf{< 0.001} & \textbf{< 0.001} & \textbf{0.02} & \textbf{0.02} & 0.97      \\
        CA-GrQc                  & \textbf{< 0.001} & \textbf{< 0.001} & 0.24          & 0.33          & 0.52      \\
        CA-HepPh                 & \textbf{< 0.001} & \textbf{< 0.001} & 0.08          & 0.08          & 0.82      \\
        CA-HepTh                 & \textbf{< 0.001} & \textbf{< 0.001} & \textbf{0.02} & \textbf{0.02} & 1.00      \\
        CA-CondMat               & \textbf{< 0.001} & \textbf{< 0.001} & \textbf{0.02} & \textbf{0.02} & 0.26      \\
        Email-Enron              & \textbf{< 0.001} & \textbf{< 0.001} & \textbf{0.02} & \textbf{0.02} & 0.12      \\
        ego-facebook             & \textbf{0.02}    & \textbf{0.001}   & \textbf{0.02} & \textbf{0.02} & 0.11      \\
        socfb-Brandeis99         & 0.09             & 0.37             & \textbf{0.02} & \textbf{0.02} & 0.38      \\
        socfb-nips-ego           & 1.00             & 1.00             & 1.00          & 1.00          & 1.00      \\
        socfb-Mich67             & 0.48             & 0.52             & \textbf{0.02} & \textbf{0.02} & 1.00      \\
        soc-gplus                & \textbf{< 0.001} & \textbf{< 0.001} & 0.16          & 0.16          & 0.37      \\
        musae\_git               & 0.48             & 0.45             & \textbf{0.02} & \textbf{0.02} & 0.97      \\
        loc-gowalla\_edges       & \textbf{< 0.001} & \textbf{< 0.001} & \textbf{0.02} & \textbf{0.02} & 0.62      \\
        gemsec\_facebook\_artist & \textbf{0.01}    & \textbf{0.02}    & \textbf{0.02} & \textbf{0.02} & 1.00      \\
        deezer\_HR               & \textbf{< 0.001} & \textbf{< 0.001} & \textbf{0.02} & \textbf{0.02} & 0.19      \\
        com-dblp                 & \textbf{< 0.001} & \textbf{< 0.001} & \textbf{0.03} & \textbf{0.03} & 0.16      \\
        Amazon0302               & \textbf{< 0.001} & \textbf{< 0.001} & \textbf{0.02} & \textbf{0.02} & 0.60      \\
        Amazon0312               & \textbf{< 0.001} & \textbf{< 0.001} & \textbf{0.02} & \textbf{0.02} & 0.33      \\
        Amazon0505               & \textbf{< 0.001} & \textbf{< 0.001} & \textbf{0.02} & \textbf{0.02} & 0.73      \\
        Amazon0601               & \textbf{< 0.001} & \textbf{< 0.001} & \textbf{0.02} & \textbf{0.02} & 0.62
    \end{tabular}
\end{table*}

\subsubsection{Comparison of the Run Time}
\label{sec:experiments:results:runTime}

\Cref{tab:runTime} shows the average run time in seconds of some of the algorithms we consider.
We note that we did not run the MMAS and the MMAS-Learn ourselves but simply report the values from \citet{SanchezSB23QLearning}.
Since their setup uses a different machine, the results are not directly comparable.
However, since we as well as \citet{SanchezSB23QLearning} ran the BRKGA, we have a common ground for comparison.
We find that our run times are close to those reported by \citet{SanchezSB23QLearning}.
Hence, we are claim that our results are in general comparable.

We note that the run time of the MMAS and the MMAS-Learn (without the time spent on learning and tuning) is faster than the run time of the fastBRKGA by a factor of around~$2$.
However, the fastBRKGA does not require any preparations upfront, and its run time is the entire time spent on this algorithms.
The MMAS variants were subject to some offline parameter tuning.
Thus, the factor of~$2$ lost in the run time comparison is a fair price to pay.

\section{Conclusion}
\label{sec:conclusion}

We proposed two ways of modifying the BRKGA~\cite{SerranoB22BRKGA}, a state-of-the-art heuristic solver for the target set selection problem (TSS).
Our first modification aims to choose the parameter values of the BRKGA during the run instead of tuning them beforehand expensively offline.
We choose the value of each parameter with respect to a power-law distribution anew in each iteration.
The resulting \emph{fastBRKGA} algorithm yields solutions that are comparable to the highly tuned BRKGA.
This shows that the expensive offline tuning used previously can be well replaced by our cheap and easy on-the-fly parameter choice.
We believe that such a parameter choice is also a good strategy for other problems than TSS.

Our second modification is adding a simple greedy heuristic specific to the TSS, namely removing vertices from the target set when this does not destroy the target set property.
When combined with our power-law parameter choice above, the resulting algorithm significantly outperforms all state-of-the-art algorithms on almost all non-easy instances.

Overall, our results show that the methods applied in the current state of the art, namely, costly parameter tuning and complex computations such as Q-learning combined with a graph convolutional network, are currently not necessary to obtain the best heuristics for TSS problems.
Optimizing good solutions greedily and choosing parameter values on the fly are already good by themselves and, in combination, significantly better than the state of the art.
We invite other researchers to give both of our approaches a try when they perform their next experiments.
Especially, the power-law parameter choice is an easy modification that is applicable to a plethora of randomized search heuristics.

\subsection*{Acknowledgments}
This research benefited from the support of the FMJH Program Gaspard Monge for optimization and operations research and their interactions with data science.

\bibliographystyle{ACM-Reference-Format.bst}
\bibliography{ich_master.bib,alles_ea_master.bib,rest.bib}

\end{document}